\documentclass[runningheads]{llncs}
\usepackage{graphicx}
\usepackage{hyperref}
\hypersetup{
colorlinks   = true,
citecolor    = blue,
urlcolor    =  blue
}

\usepackage{multirow}
\usepackage{acronym}
\usepackage{makecell}
\usepackage{hhline}
\usepackage{tabularx}
\usepackage{dcolumn}
\newcolumntype{e}[1]{D{.}{.}{#1}}
\makeatletter
\newcolumntype{f}[3]{>{\boldmath\DC@{#1}{#2}{#3}}c<{\DC@end}}
\makeatother

\usepackage{orcidlink}
\renewcommand{\orcidID}{\orcidlink}

\newcommand{\footURL}[1]{\footnote{\url{#1}}}

\makeatletter
\newcommand\footnoteref[1]{\protected@xdef\@thefnmark{\ref{#1}}\@footnotemark}
\makeatother

\begin{document}
\title{\texttt{M2DS}: Multilingual Dataset for Multi-document Summarisation}

\author{Kushan Hewapathirana\inst{1,2}\orcidID{0009-0008-1580-0699} \and
Nisansa de Silva\inst{1}\orcidID{0000-0002-5361-4810} \and
C.D. Athuraliya\inst{2}\orcidID{0009-0007-4696-5210}}
\authorrunning{K. Hewapathirana et al.}
%
\institute{Dept. of Computer Science \& Engineering, University of Moratuwa, Sri Lanka \\
\email{\{kushan.22,nisansa\}@cse.mrt.ac.lk}\and
ConscientAI, Sri Lanka\\
\email{\{kushan,cd\}@conscient.ai}}

\maketitle              
\begin{abstract}
In the rapidly evolving digital era, there is an increasing demand for concise information as individuals seek to distil key insights from various sources. Recent attention from researchers on Multi-document Summarisation (MDS) has resulted in diverse datasets covering customer reviews, academic papers, medical and legal documents, and news articles. However, the English-centric nature of these datasets has created a conspicuous void for multilingual datasets in today’s globalised digital landscape, where linguistic diversity is celebrated. Media platforms such as British Broadcasting Corporation (BBC) have disseminated news in 20+ languages for decades. With only 380 million people speaking English natively as their first language, accounting for less than 5\% of the global population, the vast majority primarily relies on other languages. These facts underscore the need for inclusivity in MDS research, utilising resources from diverse languages. Recognising this gap, we present the Multilingual Dataset for Multi-document Summarisation (M2DS), which, to the best of our knowledge, is the first dataset of its kind. It includes document-summary pairs in five languages from BBC articles published during the 2010-2023 period. This paper introduces M2DS, emphasising its unique multilingual aspect, and includes baseline scores from state-of-the-art MDS models evaluated on our dataset.

\keywords{Multi-document Summarisation  \and Multilingual \and Natural Language Processing}
\end{abstract}

\section{Introduction}
The art of document summarisation relies on intricate language skills: the ability to navigate through extensive texts, extract important information, and distil it into concise summaries. In recent years, the surge in deep learning within Natural Language Processing (NLP) has sparked significant interest among researchers in this particular task~\cite{ma2020multi,afsharizadeh2022survey,abid2022multi}. Summarisation stands as a significant challenge in NLP, gaining paramount importance as the demand for easily digestible content continues to soar~\cite{li2013multi,elhadad2013multi}.

The field of multi-document summarization~(MDS) faces a shortage of comprehensive datasets, unlike the advancements in single-document summarization~(SDS). While SDS datasets have expanded to include multilingual summarization, MDS is still relatively new but shows promise. Recent MDS research has explored various domains, such as customer reviews, academic papers, medical and legal documents, and news articles, with a predominant focus on the English language~\cite{ma2020multi,afsharizadeh2022survey,abid2022multi,li2013multi}. Despite the availability of extensive SDS datasets like CNN/Daily Mail~\cite{hermann2015teaching}, Gigaword Corpus~\cite{napoles2012annotated}, Newsroom corpus~\cite{grusky2018newsroom}, and New York Times~\cite{NYT:2008}, there is a scarcity of datasets specifically designed for versatile MDS applications, though MDS datasets like DUC\footURL{https://duc.nist.gov}, TAC\footURL{https://tac.nist.gov}, and Multi-News~\cite{fabbri2019multi} exist which predominantly serve the news domain and limited to the English language.

However, in a world boasting over 7,000 languages, the crucial requirement for multilingual approaches in MDS has become evident. Consider English, with its lexicon of over 171,146 words and a staggering 1.5 billion speakers, versus languages like Sinhala, spoken by approximately 16 million people in Sri Lanka ~\cite{simons2023ethnologue}. 
According to the 26th edition of Ethnologue published in 2023, only 380 million people speak English natively as their first language, which accounts for less than 5\% of the global population, and the total English-speaking population (i.e. as the first language and second language) is 20\% by 2023, which means that the vast majority of the global population is primarily dependent on other languages~\cite{simons2023ethnologue}. This underscores the importance of an inclusive approach in MDS research, where multilingual models cater for diverse languages. 

To address this, the research introduces the Multilingual Dataset for Multi-document Summarisation (M2DS). This dataset aims to facilitate the development of robust MDS models across diverse languages, including low-resource languages, for real-world applications. Covering languages such as English, Japanese, Korean, Tamil, and Sinhala, M2DS is considered a pioneering effort in multilingual MDS, complementing existing single-document summarisation datasets in the multilingual domain.

\section{Related Work}

This section aims to delve into the MDS landscape, exploring existing datasets, multilingual text summarisation datasets, and current state-of-the-art models. This provides insights into the diverse facets and recent advancements in MDS.

\subsection{Major MDS Datasets Across Diverse Domains}

Despite being essential for various applications, MDS datasets are relatively scarce compared to SDS datasets. However, the following key datasets have significantly influenced summarisation research. DUC and TAC datasets had set early benchmarks in the news domain~\cite{ma2020multi,afsharizadeh2022survey}. The Multi-News~\cite{fabbri2019multi} dataset offers substantial size and traceability in the news domain. WikiSum~\cite{liu2018generating} leverages Wikipedia and search engine results for abstractive summarisation challenges. Multi-XScience~\cite{lu2020multi} blends arXiv~\footURL{https://arxiv.org} papers and Microsoft Academic Graph~\cite{sinha2015overview} (MAG) for scientific writing challenges. BigSurvey~\cite{liu2023generating} and MS\^2~\cite{deyoung2021msˆ2} contribute to scientific writing, focusing on comprehensive summaries and consolidating conflicting evidence, respectively.

Domain-specific datasets like Rotten Tomatoes~\cite{Leon2020Rotten} and WikiHow~\cite{koupaee2018wikihow} diversify summarisation research into movie reviews and knowledge base articles. In customer reviews, Opinosis~\cite{ganesan2010opinosis} and OPOSUM~\cite{angelidis2018summarizing} are significant, with Opinosis providing professional-written golden summaries for model training and evaluation, and OPOSUM including domain and polarity information across six product categories.

However, a notable gap exists in these datasets—they primarily cater to English, highlighting the pressing need for multilingual datasets. Embracing linguistic diversity can drive global advancements in summarisation research, marking a fertile ground for innovation and exploration. The development of summarisation datasets in multiple languages stands as a promising avenue for future research and inclusivity in the field.

\subsection{Existing MDS Models}

Transformer architecture-based models, particularly those pre-trained on large datasets, have gained attention for their ability to capture inter-document relationships and generate informative summaries. Examples include BERTSUM~\cite{liu2019text}, using a hierarchical encoder, BART~\cite{lewis2020bart}, designed as a denoising auto-encoder, PEGASUS~\cite{zhang2020pegasus}, leveraging self-supervised learning, and T5~\cite{raffel2020exploring}, a text-to-text transformer.

In the MDS domain, PRIMERA~\cite{xiao2022primera}, based on the LongFormer Encoder-Decoder (LED)~\cite{beltagy2020longformer} architecture, stands out, surpassing previous models with a synthetic summary generation strategy during pre-training. DAMEN~\cite{moro2022discriminative}, tailored for the medical domain, combines BERT models with discriminative methods. CGSUM~\cite{chen2022comparative} introduces a citation-guided summarization approach for scientific papers.

Despite these advancements, challenges persist in accurately reflecting conflicting information, especially in multi-document scenarios~\cite{deyoung2023multi}. In multilingual MDS, progress is limited, often relying on linear programming models, and summaries are often in English rather than the original languages, limiting language coverage~\cite{marina2013multilingual}.

\subsection{Prior Work on Multilingual MDS}

The Workshop on Multilingual Summarisation (MultiLing)\footURL{https://aclanthology.org/venues/multiling/} within the ACL anthology has been a crucial focal point in multilingual summarisation research and the 2013 workshop specifically focused on Multilingual MDS~\cite{giannakopoulos2013multi}. 
During this event, a Multilingual MDS corpus was constructed, featuring languages like Arabic, English, Greek, Chinese, Romanian, Czech, Hebrew, and Spanish. The corpus creation involved selecting English texts and employing a sentence-by-sentence translation approach for the featured languages~\cite{li2013multi,elhadad2013multi}.

In terms of model concepts, Marina et al.(2013)~\cite{marina2013multilingual} introduced a novel text representation model extending the classic Vector Space Model~\cite{salton1975vector} to Hyperplane and Half-spaces. They reformulated the extractive summarisation problem as an optimisation task using linear programming, addressing the challenge of representing a large number of extracts without explicit computation. The optimal solution was found by minimising a distance function in polynomial time. While an evaluation was not conducted, the authors suggested potential assessments using Recall-Oriented Understudy for Gisting Evaluation (ROUGE) scores~\cite{lin2004rouge}. This nuanced approach to multilingual MDS, focusing on innovative text representation models and optimisation strategies, has laid a foundation for further exploration and evaluation in subsequent research endeavours~\cite{marina2013multilingual}.

\subsection{Existing Multilingual Text Summarisation Datasets}

In recent years, there has been a notable increase in research exploring the benefits of summarising across diverse languages, particularly in bilingual settings~\cite{verma2023large}. Our focus is directed towards datasets relevant to multilingual summarising, covering SDS, MDS, and Cross-Lingual Summarization (CLS).

While multilingual SDS has seen significant progress, research in multilingual MDS is limited. CLS, involving generating summaries in one language for documents in another language, has gained momentum with the development of multilingual SDS. Many multilingual SDS efforts have transitioned to include CLS components in their datasets~\cite{li2013multi,elhadad2013multi,verma2023large,hasan2021xl}.

Key multilingual SDS datasets include MLSUM~\cite{scialom2020mlsum}, featuring 1.5 million news articles across multiple languages; XL-Sum~\cite{hasan2021xl}, a diverse dataset containing 1.35 million articles in 44 languages; WikiLingua~\cite{ladhak2020wikilingua}, one of the largest parallel multilingual summarization datasets; MLGSum~ \cite{wang2021contrastive}, drawing from various news providers; and M3LS, comprising over 1.11 million multilingual multi-modal instances across 20 languages. Despite their contributions for SDS, the exploration of multilingual MDS is limited due to the absence of a dedicated high-quality dataset.

\section{M2DS Dataset}

\begin{figure*}[htbp]
    \centering
    \includegraphics[trim={3.2cm 17.07cm 4.2cm 2.91cm},clip,width=1\linewidth]{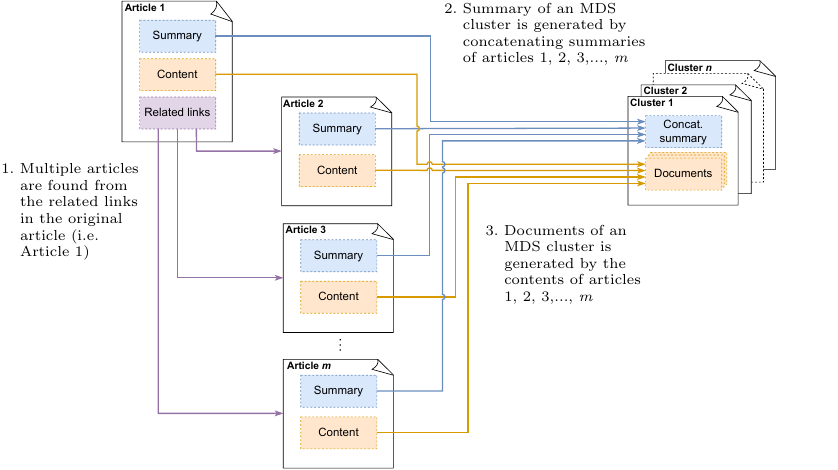}
    \caption{Process of dataset development. The golden summary for each original article was generated by logically combining its own summary and summaries of its related articles, whereas the original article and the related articles served as the collection of multi-documents. These pairs formed multi-document clusters.}
    \label{fig:Example summary 1}
\end{figure*}

This section provides an overview of the data sources, collection, and pre-processing procedures employed in this study. The dataset, named M2DS, consists of news articles in five languages, each paired with professionally written summaries sourced from the BBC. The summaries, crafted by editors, include links to the original articles for reference. The study emphasises transparency and reproducibility, with a commitment to providing links and scripts for replicating the dataset from the specified sources.

\newcolumntype{d}{S[table-format=2.3, table-number-alignment=center]}

\begin{table*}[htbp]
\centering

\resizebox{1\columnwidth}{!}{%
\begin{tabular}{|l|e{4.4}|e{4.4}|e{4.4}|l|}
\hline
\multicolumn{1}{|c}{\textbf{Dataset}} & \multicolumn{1}{|l}{\makecell{\textbf{No. of} \\ \textbf{documents}}} & \multicolumn{1}{|l}{\makecell{\textbf{No. of} \\ \textbf{clusters}}} & \multicolumn{1}{|l}{\makecell{\textbf{Avg. no. of} \\ \textbf{documents per} \\ \textbf{cluster}}} & \multicolumn{1}{|c|}{\textbf{Domain}} \\
\hline
\textbf{Multi-News\textsuperscript{\textbullet}}              & 56.0\mathrm{k}\textsuperscript{\textasteriskcentered}   &  16.0\mathrm{k} & 3.5\textsuperscript{\textasteriskcentered}  & News articles\textsuperscript{\textbullet}                                    \\ \hline
\textbf{Multi-Xscience\textsuperscript{\textopenbullet}}          & 40.0\mathrm{k}\textsuperscript{\textasteriskcentered}   & 14.0\mathrm{k} & 2.8\textsuperscript{\textasteriskcentered}  & \makecell[l]{Related work section   in \\ scientific articles\textsuperscript{\textopenbullet}}    \\ \hline
\textbf{Wikisum\textsuperscript{\textbrokenbar}}                 &  1.5\mathrm{M}\textsuperscript{\textasteriskcentered}  & 37.5\mathrm{k} & 40.0\textsuperscript{\textasteriskcentered}   & Wikipedia articles\textsuperscript{\textbrokenbar}                               \\ \hline 
\textbf{BigSurvey-MDS\textsuperscript{\textcent}}           &  430.0\mathrm{k}\textsuperscript{\textasteriskcentered}  & 7.0\mathrm{k} & 61.4\textsuperscript{\textasteriskcentered} & \makecell[l]{Human-written survey  \\ papers on various domains\textsuperscript{\textcent}} \\ \hline
 \textbf{PEERSUM\textsuperscript{\textbardbl}}                 & 11.9\mathrm{k}\textsuperscript{\textbardbl} & 1.5\mathrm{k} & 7.8\textsuperscript{\textbardbl} & \makecell[l]{Peer reviews of \\ scientific publications}        \\ \hline
\textbf{MS\textasciicircum{}2\textsuperscript{\dag}} &
  470.0\mathrm{k}\textsuperscript{\dag} & 20.0\mathrm{k} &
  23.5\textsuperscript{\dag} &
  \makecell[l]{Reviews of scientific \\  publications in medical \\ domain\textsuperscript{\dag}} \\ \hline
\makecell[l]{\textbf{Rotten Tomato} \\ \textbf{Dataset}\textsuperscript{$\uparrow$}} & 244.0\mathrm{k}\textsuperscript{\ddag}  &  9.0\mathrm{k} & 26.8\textsuperscript{\ddag} & Movie reviews\textsuperscript{\ddag}                                    \\ \hline
\textbf{M2DS} & 180.0\mathrm{k} & 51.5\mathrm{k} & 3.5 & News articles \\
~~\footnotesize{-} English & 67.0\mathrm{k}  & 17.0\mathrm{k} & 3.9 & \\ 
~~\footnotesize{-} Tamil & 32.0\mathrm{k}  & 10.0\mathrm{k} & 3.2 & \\ 
~~\footnotesize{-} Japanese & 29.0\mathrm{k}  & 11.0\mathrm{k} & 2.6 & \\
~~\footnotesize{-} Korean & 27.0\mathrm{k}  & 8.0\mathrm{k} & 3.4 & \\ 
~~\footnotesize{-} Sinhala & 23.5\mathrm{k}  & 5.5\mathrm{k} & 4.2 & \\ \hline
\end{tabular}%
}
\caption{MDS datsaset statistics. The sources are as follows: \textsuperscript{\textasteriskcentered}Xiao et al.(2022)~\cite{xiao2022primera}, \textsuperscript{\dag}DeYoung et al.(2021)~\cite{deyoung2021msˆ2}, \textsuperscript{\ddag}DeYoung et al.(2023)~\cite{deyoung2023multi}, \textsuperscript{\textbullet}Fabbri et al.(2019)~\cite{fabbri2019multi}, \textsuperscript{\textopenbullet}Lu et al.(2020)~\cite{lu2020multi}, 
\textsuperscript{\textbardbl}Li et al.(2022)~\cite{li2022peersum},
\textsuperscript{\textbrokenbar}Liu et al.(2018)~\cite{liu2018generating}, \textsuperscript{\textcent}Liu et al.(2023)~\cite{liu2023generating}, \textsuperscript{$\uparrow$}Leon et al.(2020)~\cite{Leon2020Rotten}.} 
\label{tab:dataset_mds_comparison}
\end{table*}

\subsection{Dataset Development}

In the rapidly changing digital landscape, the significant increase in online news articles has led to a growing demand for concise and informative content. To address this need, the Multilingual Multi-document Summarisation Dataset (M2DS) has been introduced. Emphasising language inclusivity, the dataset focuses on linguistic diversity and uses BBC News as the primary source due to its global coverage and articles available in multiple languages.

We utilised the M3LS dataset to extract links of parsed articles in each language. This dataset served as a valuable foundation for creating our dataset by providing corresponding Twitter page links for each BBC news article. To ensure the reliability of the M3LS dataset~\cite{verma2023large}, the authors conducted a manual assessment of article and summary quality, evaluating factors such as informativeness, length, and the ability to capture essential information. This assessment involved a meticulous review of 100 articles in four languages from their dataset. For each article, the authors carefully read the text and assigned a score between 1-5 to the golden summary, with 5 representing the best possible summary that captures most of the crucial information from the given article and vice versa. Notably, more than 70 articles across the evaluated languages received a score of over 4 out of 5 in their analysis~\cite{verma2023large}.

Assuming uniformity in the quality of articles published by BBC across various domains, the authors extrapolated that this high-quality standard holds true for every language in their dataset~\cite{verma2023large}. This verification process ensured the overall quality and reliability of the BBC articles and their summaries derived from the M3LS dataset for our study.

Once we completed the first stage, we had a SDS dataset similar to the M3LS dataset, consisting of (1) BBC articles extracted from the links included in the M3LS dataset, (2) the corresponding summaries, and (3) links to related articles which are listed in the original articles. The transformation from an SDS dataset to a MDS dataset involved extracting article-summary pairs from related links, which is illustrated by Figure \ref{fig:Example summary 1}. To ensure the quality of these summaries and the relatedness of articles in each cluster, we manually verified a sample of 10 clusters, containing 2-10 articles per cluster for English, Tamil, and Sinhala languages.

Our dataset spans from 2010 to 2021, incorporating articles sourced from the M3LS dataset dated between 2010 and 2021. To expand temporal coverage, we collected links from the front page of the BBC News site, focusing on articles from December 2021 to December 2023, and ensured non-duplication. To handle duplicates, we meticulously removed repeated links, guaranteeing the uniqueness of each document cluster. Consequently, the dataset contains a diverse range of articles, free from duplication.
The dataset is structured in the Hugging Face \verb|DatasetDict| format, offering ease of access\footnote{The dataset can be found at \url{https://huggingface.co/datasets/KushanH/m2ds} and \url{https://osf.io/7gjtm/files/osfstorage}. The code and pre-trained models are available at \url{https://github.com/KushanMH/m2ds}.}.

\subsection{Dataset Composition}
The M2DS dataset encompasses articles in Sinhala, English, Japanese, Korean, and Tamil languages. Our aspiration is for the M2DS dataset to serve as a catalyst, sparking research interest in languages that have received less exploration. Each language-specific cluster within the dataset comprises two to ten documents.

The M2DS dataset comprises 180,000 documents organized into 51,500 clusters, with an average of 3.5 documents per cluster. English-language news articles contribute the highest number of documents at 67,000, whereas Sinhala has the lowest count at 23,500. The average documents per cluster vary across languages, with Japanese news having the lowest at 2.6 and Sinhala having the highest at 4.2. (See Table~\ref{tab:dataset_mds_comparison}).

\begin{table}[tb]
\centering
\resizebox{0.9\columnwidth}{!}{%
\begin{tabular}{|c|l|e{7.7}|e{7.7}|e{7.7}|}
\hline
\textbf{Dataset}                                &     & \multicolumn{1}{|c|}{\textbf{PRIMERA}} & \multicolumn{1}{|c|}{\textbf{PEGASUS}} & \multicolumn{1}{|c|}{\textbf{LED}} \\ \hline
\multirow{3}{*}{\textbf{Multi-News}}            & \textbf{R-1} & \multicolumn{1}{f{.}{.}{-1}|}{42.0\textsuperscript{\textasteriskcentered}}  & 32.0\textsuperscript{\textasteriskcentered}             & 17.3\textsuperscript{\textasteriskcentered}          \\
                                                & \textbf{R-2} & \multicolumn{1}{f{.}{.}{-1}|}{13.6\textsuperscript{\textasteriskcentered}}    & 10.1\textsuperscript{\textasteriskcentered}             & 3.7\textsuperscript{\textasteriskcentered}           \\
                                                & \textbf{R-L} & \multicolumn{1}{f{.}{.}{-1}|}{20.8\textsuperscript{\textasteriskcentered}}    & 16.7\textsuperscript{\textasteriskcentered}             & 10.4\textsuperscript{\textasteriskcentered}          \\ \hline
\multirow{3}{*}{\textbf{Multi-Xscience}}        & \textbf{R-1} & \multicolumn{1}{f{.}{.}{-1}|}{29.1\textsuperscript{\textasteriskcentered}}    & 27.6\textsuperscript{\textasteriskcentered}             & 14.6\textsuperscript{\textasteriskcentered}          \\
                                                & \textbf{R-2} & \multicolumn{1}{f{.}{.}{-1}|}{4.6\textsuperscript{\textasteriskcentered}}     & 4.6\textsuperscript{\textasteriskcentered}              & 1.9\textsuperscript{\textasteriskcentered}           \\
                                                & \textbf{R-L} & \multicolumn{1}{f{.}{.}{-1}|}{15.7\textsuperscript{\textasteriskcentered}}    & 15.3\textsuperscript{\textasteriskcentered}             & 9.9\textsuperscript{\textasteriskcentered}           \\ \hline
\multirow{3}{*}{\textbf{WikiSum}}               & \textbf{R-1} & \multicolumn{1}{f{.}{.}{-1}|}{28.0\textsuperscript{\textasteriskcentered}}    & 24.6\textsuperscript{\textasteriskcentered}             & 10.5\textsuperscript{\textasteriskcentered}          \\
                                                & \textbf{R-2} & \multicolumn{1}{f{.}{.}{-1}|}{8.0\textsuperscript{\textasteriskcentered}}     & 5.5\textsuperscript{\textasteriskcentered}              & 2.4\textsuperscript{\textasteriskcentered}           \\
                                                & \textbf{R-L} & \multicolumn{1}{f{.}{.}{-1}|}{18.0\textsuperscript{\textasteriskcentered}}    & 15.0\textsuperscript{\textasteriskcentered}             & 8.6\textsuperscript{\textasteriskcentered}           \\ \hline
\multirow{3}{*}{\textbf{Rotten Tomatoes}}       & \textbf{R-1} & 25.4\textsuperscript{\textbullet}             & \multicolumn{1}{f{.}{.}{-1}|}{27.4\textsuperscript{\textbullet}}    & 25.6\textsuperscript{\textbullet}          \\
                                                & \textbf{R-2} & 8.4\textsuperscript{\textbullet}              & \multicolumn{1}{f{.}{.}{-1}|}{9.5\textsuperscript{\textbullet}}     & 8.0\textsuperscript{\textbullet}           \\
                                                & \textbf{R-L} & 19.8\textsuperscript{\textbullet}             & \multicolumn{1}{f{.}{.}{-1}|}{21.1\textsuperscript{\textbullet}}    & 19.6\textsuperscript{\textbullet}\\         
\hline
\end{tabular}%
}
\caption{\label{tab:Model_performences}ROUGE scores of selected models on different domain datasets. Note: This study utilises multiple datasets from various domains.(Results obtained from Hewapathirana et al.(2023)~\cite{hewapathirana2023multi}). The Multi-News dataset~\cite{fabbri2019multi} consists of news articles, Multi-XScience~\cite{lu2020multi} focuses on scientific papers, WikiSum~\cite{liu2018generating} provides Wikipedia summaries, and Rotten Tomatoes~\cite{Leon2020Rotten} covers movie reviews. These diverse datasets offer valuable resources for training and evaluating summarization models. The sources for the results are as follows: \textsuperscript{\textasteriskcentered}Xiao et al.(2022)~\cite{xiao2022primera} and \textsuperscript{\textbullet}DeYoung et al.(2023)~\cite{deyoung2023multi}}

\end{table}

\subsection{Dataset Comparison}
As the M2DS dataset is the first of its kind, we decided to compare it with existing MDS datasets.
We conducted a comprehensive comparison with existing MDS datasets across various domains, considering factors such as the total number of documents, number of clusters, and the number of documents per cluster. This approach provides valuable insights into the positioning of our dataset within the landscape of English-centric MDS datasets.

To offer a holistic view, we present both aggregated numbers and language-specific statistics. This multi-faceted analysis allows for a nuanced understanding of the dataset's characteristics in comparison to established datasets as shown in Table \ref{tab:dataset_mds_comparison}. Comparatively, when assessing M2DS against other MDS datasets, our dataset stands out with a significant overall number of documents. However, on a language-wise comparison, it exhibits a relatively lower count per language, emphasising the importance of considering linguistic variations in dataset analysis.

It is important to note that certain statistical metrics, such as average sentence length, average token count, and average word count per article and per cluster, were not included in the comparison. The rationale behind this omission is the inherent linguistic differences across languages. For instance, languages like Japanese and Korean may convey the same meaning with a lesser number of words, or in some cases, they might encapsulate an entire sentence with a single character. Consequently, direct comparisons based on these metrics could be misleading due to the diverse linguistic structures and expressions employed by different languages.

\section{Experiments}

In this section, we outline the experiments conducted, taking the dataset size relative to existing English-centric MDS datasets into consideration. The dataset was partitioned into training, testing, and validation sets, following a 90-5-5 split for languages other than English~\cite{verma2023large}. For English, we adopted an 80-10-10 split, aligning with the practices of previous researchers in MDS dataset creation~\cite{liu2023generating,liu2018generating,koupaee2018wikihow}.
A meticulous evaluation of various MDS models was carried out to establish robust baselines. Additionally, we explored the efficacy of open-source large language models, aiming to set a strong baseline for future research.

\subsection{Pre-trained Model Selection}

In the context of Multilingual MDS, there is a noticeable gap in the literature concerning the absence of transformer-based models. Recognising the robustness of such models, the approach in this study involved evaluating pre-trained models to establish baselines. Model selection was guided by an extensive literature review, considering factors like model performance, ROUGE scores, publication year, and venue.

The evaluation focused on three summarising models: PRIMERA, PEGASUS, and LED. PRIMERA demonstrated superior performance in previous studies, while PEGASUS showed superior sentiment understanding, particularly on the Rotten Tomatoes dataset. LED, a widely used pre-trained model, served as a baseline in the existing literature (See Table \ref{tab:Model_performences}).

Among the models under consideration, PRIMERA emerged as a promising choice due to its distinctive approach to MDS~\cite{hewapathirana2023multi}. Seeking to minimize dependency on dataset-specific modeling, it consolidates multiple documents into a single extended sequence, employing the LED architecture known for its computational efficiency. PRIMERA incorporates a sparse ``local+global" attention mechanism in the encoder and introduces special document separator tokens (\verb|<doc-sep>|) to indicate document boundaries. Inspired by models like PEGASUS, PRIMERA adopts a unique masking strategy based on the Entity Pyramid framework, to address the limitations in selecting representative information for summarisation~\cite{xiao2022primera,lewis2020bart,beltagy2020longformer}.

\begin{table*}[!htbp]
\centering
\resizebox{1\columnwidth}{!}{%
\begin{tabular}{|c|c|*{7}{e{5.5}|}}
\hline
\multirow{2}{*}{\textbf{Language}} &
  \multirow{2}{*}{\textbf{}} &
  \multicolumn{7}{c|}{\textbf{Models}} \\ \cline{3-9} 
 &
   &
  \multicolumn{1}{c|}{\textbf{LEAD-3}} &
  \multicolumn{1}{c|}{\textbf{RANDOM}} &
  \multicolumn{1}{c|}{\textbf{CENTROID}} &
  \multicolumn{1}{c|}{\textbf{PRIMERA}} &
  \multicolumn{1}{c|}{\textbf{PEGASUS}} &
  \multicolumn{1}{c|}{\textbf{LED}} &
  \multicolumn{1}{c|}{\textbf{Llama 2}} \\ \hline
\multirow{3}{*}{\textbf{Sinhala}} &
  \textbf{R-1} &
  0.06 &
  5.7 &
  4.5 &
  5.7 &
 4.1 &
 3.6 & 
 \multicolumn{1}{f{.}{.}{-1}|}{20.2}
   \\ \cline{2-9} 
 &
  \textbf{R-2} &
  0.0 &
  0.05 &
  0.1 &
  2.2 &
  2.1 &
  1.9 &
  \multicolumn{1}{f{.}{.}{-1}|}{6.5}
   \\ \cline{2-9} 
 &
  \textbf{R-L} &
  0.06 &
  5.1 &
  3.9 &
  3.2 &
  2.8 &
  2.9 &
  \multicolumn{1}{f{.}{.}{-1}|}{17.3}
   \\ \hline
\multirow{3}{*}{\textbf{Japanese}} &
  \textbf{R-1} &
  3.5 &
  2.3 &
  1.9 &
  6.3 &
  5.7 &
  5.9 &
  \multicolumn{1}{f{.}{.}{-1}|}{7.7}
   \\ \cline{2-9} 
 &
  \textbf{R-2} &
  0.0 &
  0.01 &
  0.05 &
  \multicolumn{1}{f{.}{.}{-1}|}{3.2} &
  1.3 &
  1.4 &
  0.8 
   \\ \cline{2-9} 
 &
  \textbf{R-L} &
  3.5 &
  1.9 &
  1.7 &
  4.1 &
  3.3 &
  2.7 &
  \multicolumn{1}{f{.}{.}{-1}|}{6.8}
   \\ \hline
\multirow{3}{*}{\textbf{Korean}} &
  \textbf{R-1} &
  2.4 &
  1.4 &
  1.3 &
  5.4 &
  5.5 &
  4.6 &
  \multicolumn{1}{f{.}{.}{-1}|}{8.5}
   \\ \cline{2-9} 
 &
  \textbf{R-2} &
  0.4 &
  0.02 &
  0.03 &
  1.1 &
  \multicolumn{1}{f{.}{.}{-1}|}{1.4} &
  0.8 &
  1.0
   \\ \cline{2-9} 
 &
  \textbf{R-L} &
  2.3 &
  1.3 &
  1.3 &
  2.3 &
  2.9 &
  1.9 &
  \multicolumn{1}{f{.}{.}{-1}|}{8.1}
   \\ \hline
\multirow{3}{*}{\textbf{Tamil}} &
  \textbf{R-1} &
  6.8 &
  1.6 &
  2.2 &
  4.4 &
  3.8 &
  3.7 &
  \multicolumn{1}{f{.}{.}{-1}|}{10.2}
   \\ \cline{2-9} 
 &
  \textbf{R-2} &
  0.9 &
  0.0 &
  0.06 &
  1.1 &
  0.7 &
  0.4 &
  \multicolumn{1}{f{.}{.}{-1}|}{3.1}
   \\ \cline{2-9} 
 &
  \textbf{R-L} &
  6.2 &
  1.6 &
  1.9 &
  2.2 &
  1.7 &
  1.3 &
  \multicolumn{1}{f{.}{.}{-1}|}{9.8}
   \\ \hline
\multirow{3}{*}{\textbf{English}} &
  \textbf{R-1} &
  1.2 &
  6.4 &
  7.6 &
  \multicolumn{1}{f{.}{.}{-1}|}{28.7} &
  22.5 &
  20.5 &
  20.8
   \\ \cline{2-9} 
 &
  \textbf{R-2} &
  0.0 &
  0.05 &
  3.8 &
  12.3 &
  9.9 &
  10.1 &
  \multicolumn{1}{f{.}{.}{-1}|}{13.5}
   \\ \cline{2-9} 
 &
  \textbf{R-L} &
  1.1 &
  5.7 &
  7.6 &
  17.1 &
  14.7 &
  15.2 &
  \multicolumn{1}{f{.}{.}{-1}|}{19.2}
   \\ \hline
\end{tabular}%
}
\caption{Comparison of performance across fine-tuned models on the M2DS dataset }
\label{tab:results-table}
\end{table*}

\subsection{Baselines}
For our baseline models, we explore simpler extractive approaches and statistical methods alongside pretrained models. In the extractive category, we employ LEAD-3 and RANDOM~\cite{verma2023large}. LEAD-3 extracts the first three sentences from the source text as the final summary, while RANDOM recursively selects words randomly from the source text until the threshold summary length is reached. These approaches serve as unbiased reference points for understanding and comparing more complex models.
In the statistical approach, we experiment with CENTROID, inspired by \cite{radev2000centroid}. CENTROID ranks sentences based on centrality scores derived from the words within each sentence, utilising TF-IDF scores to measure word similarity. We extract top sentences from each ranking until the threshold summary length is achieved.

Moving to pre-trained models, we select PRIMERA, PEGASUS, and LED, training them on each language's respective training set. For tokenization, we use a space-based tokenizer for Sinhala and Tamil, the original tokenizer for PRIMERA and LED in other languages, and a space-based tokenizer for PEGASUS in all languages except English. For English, we report results both with and without fine-tuning.

Additionally, we present baseline scores for Llama 2, which is an open Large Language Model (LLM) ~\cite{touvron2023llama}. Llama 2, an updated version of Llama 1 and a formidable 7 billion-parameter causal decoder-only model, is introduced by Meta AI~\footURL{https://ai.meta.com}. 
We limit ourselves to using open LLMs to ensure reproducibility within the research community.

\begin{table*}[htbp]
\centering
\resizebox{1\columnwidth}{!}{%
\begin{tabular}{|c|l|e{3.3}|e{3.3}|e{3.3}|e{3.3}|e{3.3}|e{3.3}|}
\hline
\multirow{2}{*}{\textbf{Language}} &
  \multirow{2}{*}{} &
  \multicolumn{6}{c|}{\textbf{Models}} \\ \cline{3-8} 
 &
   &
  \multicolumn{1}{l|}{\textbf{PRIMERA}} &
  \multicolumn{1}{l|}{\makecell{\textbf{PRIMERA} \\ (fine-tuned)}} &
  \multicolumn{1}{l|}{\textbf{PEGASUS}} &
  \multicolumn{1}{l|}{\makecell{\textbf{PEGASUS} \\ (fine-tuned)}} &
  \multicolumn{1}{l|}{\textbf{LED}} &
  \multicolumn{1}{l|}{\makecell{\textbf{LED} \\ (fine-tuned)}} \\ \hline
\multirow{3}{*}{\textbf{English}} &
  \textbf{R-1} &
  23.6 &
  \multicolumn{1}{c|}{\textbf{28.7}} &
  18.6 &
  22.5 &
  17.1 &
  20.5
   \\ \cline{2-8} 
 &
  \textbf{R-2} &
  8.8 &
  \multicolumn{1}{c|}{\textbf{12.3}} &
  9.1 &
  9.9 &
  7.1 &
  10.1
   \\ \cline{2-8} 
 &
  \textbf{R-L} &
  13.6 &
  \multicolumn{1}{c|}{\textbf{17.1}} &
  12.4 &
  14.7 &
  13.2 &
  15.2
   \\ \hline
\end{tabular}%
}
\caption{Comparison of performance across models originally trained on English datasets, on English articles of the M2DS dataset}
\label{tab:fine-tune-table}
\end{table*}

\section{Analysis and Discussion}

In our baseline evaluations, Llama 2 7B outperforms all other models, showcasing its robust performance. Notably, PRIMERA excels slightly better in the English language, indicating its effectiveness in capturing linguistic nuances specific to that language. However, when assessing the state-of-the-art MDS models fine-tuned on our dataset, we observed a discernible drop in performance compared to their previous performance under English-centric news domain datasets as depicted in Table \ref{tab:Model_performences}. This phenomenon could stem from the models struggling to capture language-specific information unique to each language in our multilingual dataset (See Table \ref{tab:results-table}).

A noteworthy observation is the lower scores in LEAD-3, which extracts only the first three sentences as the summary. This suggests that our dataset exhibits better quality, addressing the issues found in TAC/DUC datasets, where the first three sentences often serve as summaries, leading models to learn biased patterns.

Contrary to the trend of using LLMs for MDS, our findings suggest that simpler models, such as PRIMERA specifically designed for MDS tasks, may be more effective. This is evident from PRIMERA's superior performance in English when compared to Llama 2. Designing task-specific models like PRIMERA which perform well without extensive fine-tuning, could be a more effective approach with respect to resource constraints. Additionally, it is crucial to note that Llama 2, without fine-tuning, achieves competitive results, highlighting its potential for zero-shot learning and its effectiveness across diverse datasets compared to models specifically trained on individual datasets.

Furthermore, it is essential to emphasise the scalability of models like Llama 2, indicating their potential for handling larger datasets and their adaptability across various domains. Additionally, future research should explore Transfer Learning techniques to enhance the performance of MDS models across different languages, minimising the observed drop in performance. Finally, understanding the impact of dataset quality on model evaluation is crucial, and our dataset's higher quality, as reflected in low LEAD-3 scores, underscores the significance of curating datasets that truly represent the summarisation task.

Additionally, we conducted a comparison of PRIMERA's and other models' performance with and without fine-tuning on the English language subset of our dataset (See Table \ref{tab:fine-tune-table}). Although PRIMERA excels with a zero-shot approach surpassing other models, its scores are slightly lower when compared to other MDS models trained on news domain datasets. This suggests that our dataset presents challenges for models, underscoring its quality. Furthermore, all the models have improved their performance when they are fine-tuned. For instance, PRIMERA's score increased from 23.6 to 28.7, exhibiting the highest improvement among other models.

\section{Conclusion and Future Directions}

The study introduces the M2DS dataset to fill the gap in multilingual MDS datasets. While existing MDS datasets have made strides in various domains, they mostly focus on English, leaving a void in multilingual representation. M2DS, with document-summary pairs in five languages, stands out as the pioneering multilingual MDS dataset.

The evaluation of M2DS against existing datasets demonstrates its potential and unique contribution to the field. Baseline scores from state-of-the-art MDS techniques provide a benchmark for future research in multilingual settings. Llama 2 7B outperforms other models, showcasing robust performance. PRIMERA excels slightly better in English, indicating effectiveness in capturing language-specific nuances.

The introduction of M2DS opens avenues for future research, enabling researchers to enhance the robustness of MDS models across diverse linguistic contexts. Possible directions include language-specific model tuning, exploring multilingual model development, and extending M2DS into diverse domains beyond news articles for broader applications.

%
%

\bibliographystyle{splncs04}
\bibliography{bibliography}

\end{document}